\documentclass[10pt,twocolumn,letterpaper]{article}

\usepackage{cvpr}
\usepackage{times}
\usepackage{epsfig}
\usepackage{graphicx}
\usepackage{amsmath}
\usepackage{amssymb}
\usepackage{placeins}
\usepackage{float}

\usepackage[pagebackref=true,breaklinks=true,letterpaper=true,colorlinks,bookmarks=false]{hyperref}

\ifcvprfinal\fi
\begin{document}

\title{Deblurring using Analysis-Synthesis Networks Pair}

\author{
\begin{tabular}{c}
Adam Kaufman \\
{\tt\small adam.kaufman@mail.huji.ac.il}
\end{tabular}
\qquad 
\begin{tabular}{c}
Raanan Fattal \\
 {\tt\small raananf@cs.huji.ac.il}
\end{tabular}
\\
School of Computer Science and Engineering\\
The Hebrew University of Jerusalem, Israel 
}




\maketitle

\begin{abstract}
   Blind image deblurring remains a challenging problem for modern artificial neural networks. Unlike other image restoration problems, deblurring networks fail behind the performance of existing deblurring algorithms in case of uniform and 3D blur models. This follows from the diverse and profound effect that the unknown blur-kernel has on the deblurring operator. 
   
We propose a new architecture which breaks the deblurring network into an analysis network which estimates the blur, and a synthesis network that uses this kernel to deblur the image. Unlike existing deblurring networks, this design allows us to explicitly incorporate the blur-kernel in the network's training. 

In addition, we introduce new cross-correlation layers that allow better blur estimations, as well as unique components that allow the estimate blur to control the action of the synthesis deblurring action.

Evaluating the new approach over established benchmark datasets shows its ability to achieve state-of-the-art deblurring accuracy on various tests, as well as offer a major speedup in runtime. 
\end{abstract}

\section{Introduction}
\label{sec:intro}

When taking a photo using a handheld device, such as a smartphone, camera shakes are hard to avoid. If the scene is not very bright, the movement during exposure results in a blurry image. This becomes worse as the scene is darker and a longer exposure time is needed.

The task of recovering a blur-free image given a single blurry photo is called deblurring and is typically divided into blind and non-blind cases, depending on whether the blur-kernel is known and unknown respectively. Both cases were studied extensively in the computer vision literature, and were addressed using dedicated algorithms~\cite{Levin11,Shan08,Krishnan09,Xu13,Pan16,Delbracio15} and using Artificial Neural Networks (ANNs), which are either used for implementing parts of the deblurring pipeline~\cite{Schuler13, Xu14DNN, Zhang16,Chakrabarti16, Sun15} or carrying-out the entire pipeline~\cite{Nah17,Tao18,Kupyn17,Kupyn19, Nimisha17}. Due to the ill-posedness of the blind deblurring case, classic methods use various constraints and image priors to regularize the space of solutions~\cite{Fergus06,Cho09,Levin11,whyte12,Goldstein12}. Typically these methods consist of a time-consuming iterative optimization, where both the sharp image and blur-kernel are recovered, as well as relay on specific image priors that may fail on certain types of images.

In these respects, the ANN-based solutions offer a clear advantage. While they may require a long training time, their subsequent application consists of a rather efficient feed-forward procedure. Furthermore, using suitable training sets they achieve a (locally) optimal performance over wide classes of images. It is worth noting that in this domain the training sets are produced automatically by applying a set of blur-kernels over a set of sharp images involving no manual effort.

\begin{figure*}[t]
\begin{center}

\begin{tabular}
{p{0.175\linewidth}p{0.175\linewidth}p{0.175\linewidth }p{0.175\linewidth }p{0.175\linewidth }}
\begin{center}\textbf{Input} \end{center} & 
\begin{center}\textbf{No Analysis} \end{center} & \begin{center}\textbf{Ours (Analysis+Synthesis)} \end{center} & 
\begin{center}\textbf{Synthesis + GT kernel} \end{center} & 
\begin{center}\textbf{Kernel-Expert Syntheses}  \end{center} 
\end{tabular}
\vspace{-4mm}

\includegraphics[width=1\linewidth]{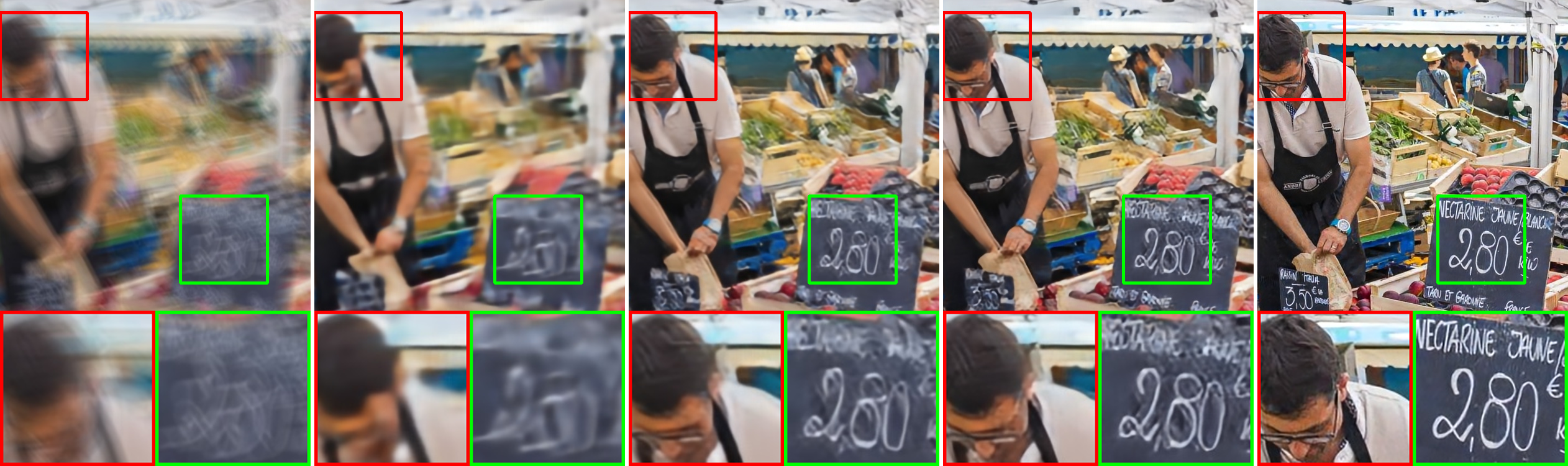} \\

  \caption{
Different network configurations. Column 2 from the left shows the results obtained when training our synthesis network \emph{without} the analysis. Like existing deblurring-networks, this requires it cope with every possible blur-kernel. The results of our network pair archives a better deblurring in Column 3. Providing the ground-truth (GT) kernel to the synthesis network achieves better results (Column 4) suggesting that our synthesis network performs better than our analysis. Finally, the result of a network trained for a single specific kernel is shown in Column 5, offering a moderate improvement. These configurations were tested on 1000 images and resulted in the following mean PSNR: No analysis 23.7, our 26.6, GT kernel 28.28, Kernel-Expert 33.75. }
\label{fig:specialized_vs_general}
\end{center}
\end{figure*}

Nevertheless, unlike other image restoration problems, such as denoising~\cite{Zhang17, Zhang18}, upscaling~\cite{Dong15,Kim16,Ledig17}, and in-painting~\cite{Pathak16, Yeh17, Yu18}, even the case of spatially-uniform blind deblurring ANN-based approaches fail to show a clear improvement over existing algorithms when it comes to the accuracy of the deblurred image. This major shortcoming stems from the fact that the space of blur-kernels, which specifies the blur degradation, is very large and has a very diverse effect over the inverse operator. Indeed, this is in contrast to the one-dimensional space of noise amplitude $\sigma$ that governs denoising. While it is shown that a single network trained over a wide range of $\sigma$ can achieve a denoising accuracy over a specific $\sigma^*$ which is comparable to a network trained over that particular value~\cite{Zhang17}. The analog experiment shows that this is far from being the case for image deblurring, as reported in~\cite{Schuler13} and demonstrated in Figure~\ref{fig:specialized_vs_general}.

In this paper we describe a novel network architecture and training scheme that account for the unique nature of the deblurring operation. The design of this new network incorporates various components found in image deblurring algorithms, and allows it to mimic and generalize their operation. More specifically, similarly to blind-deblurring algorithms that recover the blur-kernel, our architecture is divided into a pair of \emph{analysis} and \emph{synthesis} networks, where the first is trained to estimate the blur-kernel, and the second uses it to deblur the image. Unlike existing ANN-based architectures, this data path allows our training process to \emph{explicitly} utilize the ground-truth blur-kernel in the training process.

Moreover, inspired by the success of recent algorithms to recover the blur-kernel from irregularities in the auto-correlation of the blurry image, we introduce this computation as a layer of the analysis network, allowing it to perform this recovery as well as generalize it.

Finally, the recovered blur-kernel governs the non-blind deblurring operation globally and uniformly across the image. This further motivates us to introduce an additional component into the synthesis network, allowing it to encode and spread the kernel globally across its layers.

Evaluation of this new approach over the case of spatially-uniform blur (convolution) consistently demonstrates its ability to achieve superior deblurring accuracy compared to existing deblurring networks. These tests also show that the new network compares favorably to existing state-of-the-art deblurring algorithms.

\section{Previous Work}
\label{sec:prev}

Image deblurring has been the topic of extensive research in the past several decades. Since this contribution relates to the blind image deblurring, we will limit the survey of existing art to this case.

In case of 2D camera rotation, the blur is uniform across the image and can be expressed by a convolution with a single 2D blur-kernel. To resolve the ambiguity between the blur and the image inherent to the blind deblurring case, as well as to compensate for the associated data loss, different priors are used for the recovery of the latent image and blur-kernel. The image priors typically model the gradients distribution found in natural images, and the blur-kernel is often assumed to be sparse. Chan and Wong~\cite{ChanTV98} use the total variation norm for image regularization. Fergus \etal~\cite{Fergus06} employ a maximum a-posteriori estimation that uses a mixture of Gaussians to model the distribution the image gradients and a mixture of exponential distributions to model the blue-kernel. Shan \etal~~\cite{Shan08} suggest a different approximation for the heavy-tailed gradient distribution and add a regularization term to promote sparsity in the blur-kernel. Cho and Lee~\cite{Cho09} accelerates the estimation by considering the strong edges in the image. Xu and Jia~\cite{Xu10} use edge selection mask to ignore small structures that undermine the kernel estimation. Following this work, Xu \etal~\cite{Xu13} add an $L_0$ regularization term to suppress small-amplitude structures in the image to improve the kernel estimation.

Another line of works exploit regularities that natural images exhibit in the frequency domain. Yitzhaky \etal~\cite{Yitzhaky98} assume a 1D motion blur and use auto-correlation over the image derivative to find the motion direction. Hu \etal~\cite{Hu2012} use 2D correlation to recover 2D kernels and use the eight-point Laplacian for whitening the image spectrum. Goldstein and Fattal~\cite{Goldstein12} use an improved spectral whitening for recovering the correlations in the images as well as use a refined power-law model the spectrum. These method resolve the power spectrum of the kernel, and use a phase retrieval algorithm to recover its phase~\cite{Fienup82}. In order to allow our analysis network to use such a kernel recovery process, we augment it with special layers that compute the correlation between different activations.


The case of a 3D camera rotation, the blur is no longer uniform across the image but can be expressed by a 3D blur-kernel. Gupta \etal~\cite{Gupta10} as well as Whyte \etal~\cite{whyte12} use the maximum a-posteriori estimation formulation to recover these kernels. Finally, the most general case of blur arise from camera motion and parallax or moving objects in the scene, in which case a different blur applies to different objects in the scene, and is addressed by segmenting the image~\cite{Levin06}.

More recently the training and use of artificial neural networks, in particular Convolutions Neural Networks (CNNs), became the dominant approach for tackling various image restoration tasks, including image deblurring. Earlier works focused on the the non-blind uniform deblurring, for example Schuler \etal~\cite{Schuler13} use a multi-layer perceptron in order to remove the artifacts produced by a non-blind deblurring algorithm. Xu \etal~\cite{Xu14DNN} based their network on the kernel separability theorem and used the SVD of the pseudo-inverse of the blur kernel to initialized the weights of large 1D convolutions. Both those methods need to be trained per kernel, which burdens their use at run-time.

Zhang \etal~\cite{Zhang16} managed to avoid per-kernel training by first applying a non ANN based deconvolution module, followed by a FCNN to remove noise and artifacts from the gradients. This is done iteratively where the denoised gradients are used as image priors to guide the image deconvolution in the next iteration. However, this method is still limited to the non-blind case for the initial deblurring step. On the other hand, a few architecture were purposed for blur estimation. Chakrabarti~\cite{Chakrabarti16} train a network to estimate the Fourier coefficients of a deconvolution filter, and used the non-blind EPLL~\cite{Zoran11} method to restore the sharp image and generate comparable performance to state-of-the-art iterative blind deblurring methods. Sun \etal~\cite{Sun15} trained a classification network to generate a motion field for non-uniform deblurring. The classification is performed locally and restricted to 73 simple linear blur-kernels, and therefore limited to simple blurs. Both of the above methods \cite{Chakrabarti16,Sun15} use a traditional non-blind method to deblur the image.

Most recent effort focuses on end-to-end (E2E) networks, trained over dataset containing non-uniform parallax deblurring (dynamic scene). Nah \etal~\cite{Nah17} proposed an E2E multi-scale CNN which deblurs the image in a coarse-to-fine fashion. This network tries to recover the latent sharp image at different scales using a loss per level. Following this work, Tao \etal \cite{Tao18} used a scale-recurrent network with share parameters across scales to perform deblurring in different scales. This reduce the number of parameters and increase stability. Kupyn \etal \cite{Kupyn17} use a discriminator-based loss, which encourages the network to produce realistic-looking content in the deblurred image. While capable of deblurring complex scenes with multiple moving objects, those methods fall short in the uniform blur case. As we show below, this case greatly benefits from using the ground-truth kernels in the training. Nimisha \etal~\cite{Nimisha17} use an auto-encoder to deblur images, by mapping their latent representation into a blur-free one using a dedicated generator network. While this approach provides an E2E network, it was trained over a uniform blur.

\section{Method}
\label{sec:method}

The construction of the new deblurring networks pair that we describe here is inspired by the computation pipeline of existing blind deblurring algorithms~\cite{Goldstein12,Cho09}. Similarly to these algorithms our deblurring process first recovers the blur-kernel using a dedicated \emph{analysis} network. The estimated kernel, along with the input image, are then fed into a second \emph{synthesis} network that performs the non-blind deblurring and recovers the sharp image. This splitting of the process allows us to employ a blur-kernel reconstruction loss over the analysis network, on top of the sharp image reconstruction loss that applies to both networks. In Section~\ref{sec:training} we describe this training process.

Several deblurring algorithms~\cite{Yitzhaky98,Goldstein12,Hu2012} successfully recover the blur-kernel based on the irregularities it introduces to the auto-correlation of the input image. In order to utilize this computational path in our analysis network, we introduce in Section~\ref{sec:analysis} a unique layer that computes these correlations as well as generalizes them.

Finally, in case of the uniform 2D blur, i.e., convolution, that we consider (as well as the 3D non-uniform case), the sharp image recovery consists of a global operator (deconvolution) whose operation is dictated by the blur-kernel across the entire image. This fundamental fact inspires another design novelty that we introduce in the synthesis network architecture. As we describe in Section~\ref{sec:synthesis}, the synthesis network consists of a fairly-standard U-Net architecture which carries out the image sharpening operation. In order to allow the estimated kernel to dictate its operation, and do so in a global manner, we augment this network with specialized components that encode the kernel into multipliers and biases that affect the U-Net action across the entire image domain and all the layers.

The derivation of our new approach, as well as its evaluation in Section~\ref{sec:results}, assumes a uniform blur (2D convolution). In Section~\ref{sec:discussion} we discuss the option of generalizing this approach to handle more general blurring.


\begin{figure}[t]
\includegraphics[width=1\linewidth]{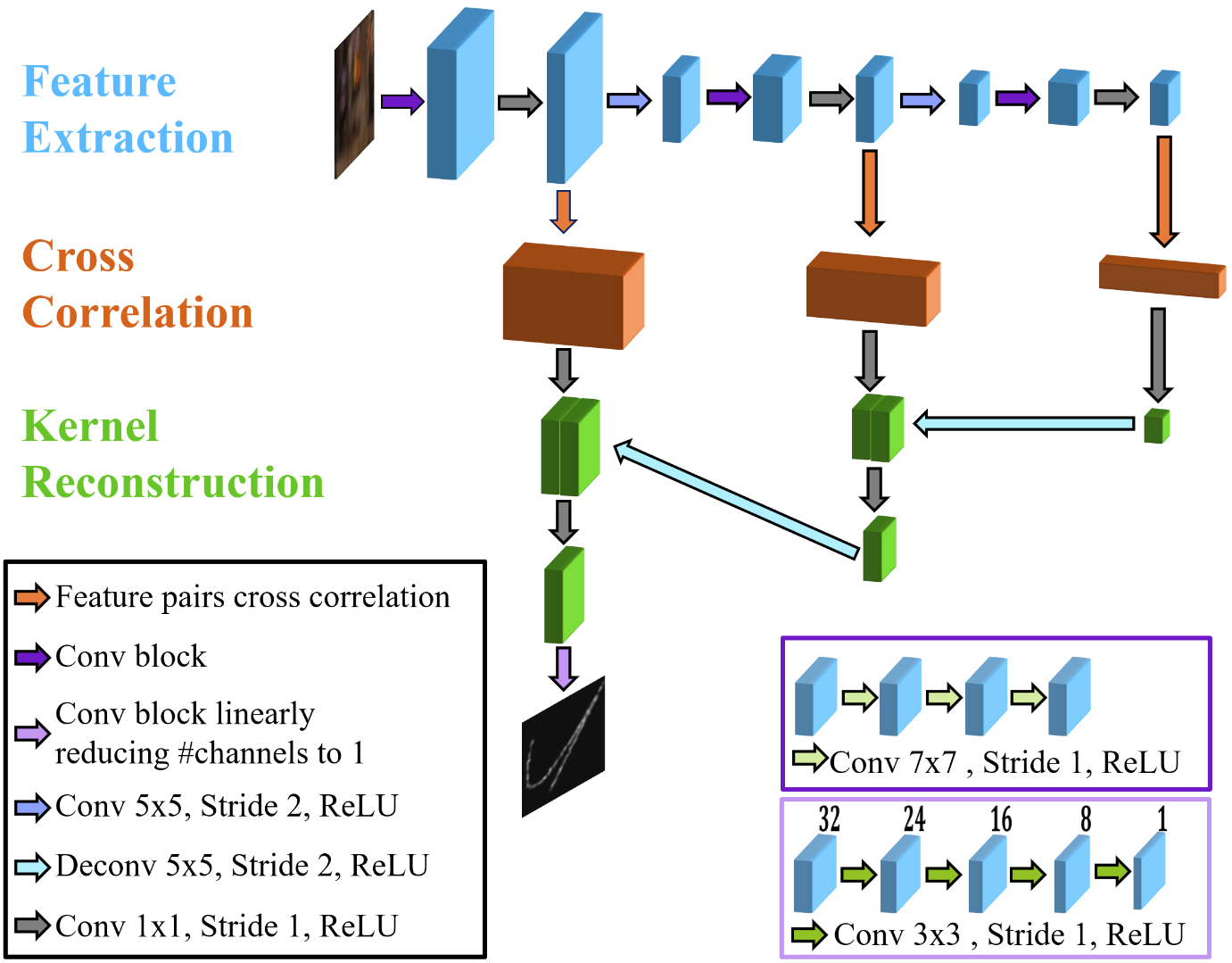}

  \caption{Analysis Network Architecture. The first stage consists of extracting features (activations) at multiple scales by applying convolution layers, and pooling operations. At the second stage, the cross-correlation between the resulting activations are computed at all scales. Finally, the estimated blur-kernel is reconstructed from coarse to fine by means of un-pooling and convolution steps. The pooling and up-pooling operations apply x2 scaling. We use 64 filters in the feature extracting step, and reduce them to 32 channels before computing the cross-correlation stage. This illustration shows only two spatial scalings, whereas our implementation consists of three.}
\label{fig:analysis_arch}
\end{figure}

\subsection{Analysis Network}
\label{sec:analysis}

The first network in our pipeline is the \emph{analysis} network which receives the input blurry image and estimates the underlining blur-kernel. The novel layers that we introduce to its architecture are inspired by previous works~\cite{Yitzhaky98,Goldstein12,Hu2012} that describe how the Fourier power-amplitudes components of the blur-kernel can be estimated from the auto-correlation function of the blurry image. These methods rely on the observation that natural images lose their spatial correlation upon differentiation. Thus, any deviations from a delta auto-correlation function can be attributed to the blur-kernel's auto-correlation (or power-amplitudes in Fourier space). 

Standard convolutional networks do not contain this multiplicative step between its filter responses, and hence mimicking this kernel-recovery pipeline may be difficult to be learned. Rather than increasing the networks capacity by means of increasing its depth, number of filters and non-linearities, we incorporate these computational elements as a new correlations layers that we add to its architecture.

\bf{Correlations Layers. }\normalfont 
The analysis network consists of the following three functional layers: (i) a standard convolutional layer that receives the input image and extracts (learnable) features from it. This process is applied recursively to produce three spatial levels. Specifically, at each level we use 3 convolution layers applying 64 filters of size 7-by-7, separated by x2 pooling, to extract feature maps for the correlation layer that follows. By convolution layers we imply also the addition of bias terms and a point-wise application of a ReLU activation function. Since the correlation between every pair of filters is quadratic in their number, we reduce their number by half (at each level), by applying 32 filters of size 1-by-1. Next, (ii) we compute the cross-correlation, $C_{ij}(s,t) = \sum_{x,y} f_i(x-s,y-t)f_j(x,y)$, between each pair of these activation maps $f_i$ at a limited spatial range of $2^{-l}m \leq s,t \leq 2^{-l}m$ pixels, where $m$ is the spatial dimension of the recovered kernel grid, and $l=0...$ is the level (scale). Note that since $C_{ij}(s,t)=C_{ji}(-s,-t)$, only half of these coefficients needs to be computed. This stage results with a fairly large number of channels, i.e., $32 \times 31=992$, which we reduce back to 32 again, using 32 filters of size 1-by-1. Note that these correlations are computed between activations derived from the same data. Hence the standard auto-correlation function can be carried out if the filters applied, at previous convolution layer, correspond to delta-functions. In this case, the correlation within each channel $C_{ii}$ will correspond to the auto-correlation of that channel. Note that by learning different filters, the network generalizes this operation.

Finally, (iii) the maps of size $(2^{-l}{m+1})$-by-$(2^{-l}{m+1})$-by-32, extracted from the correlations at every scales $l$, and are integrated back to the finest scale. This is done by recursively applying x2 un-pooling (up-sampling), followed by a convolution with 32 filters of size $5$-by-$5$. This result is concatenated with the map of the following level, and the number of channels is reduced from 64 back to 32, by applying a convolution with 32 filters of size 1-by-1.

At the finest level, the 32 channels are reduced gradually down to a single-channeled 2D blur kernel. This is done by a sequence of 3-by-3 convolution layers that produce intermediate maps of 24, 16, 8 and 1 channels. Figure~\ref{fig:analysis_arch} sketches the architecture of the analysis network. 
The resulting $m$-by-$m$ map is normalized to sum 1 in order to provide an admissible blur-kernel estimate. We use an $L_1$ loss between the true and estimated kernels when training this network. Finally, we note that the analysis network is trained to produce a monochromatic blur-kernel by considering the Y channel of the input image in YUV color-space.

\begin{figure*}[h]
\includegraphics[width=1\linewidth]{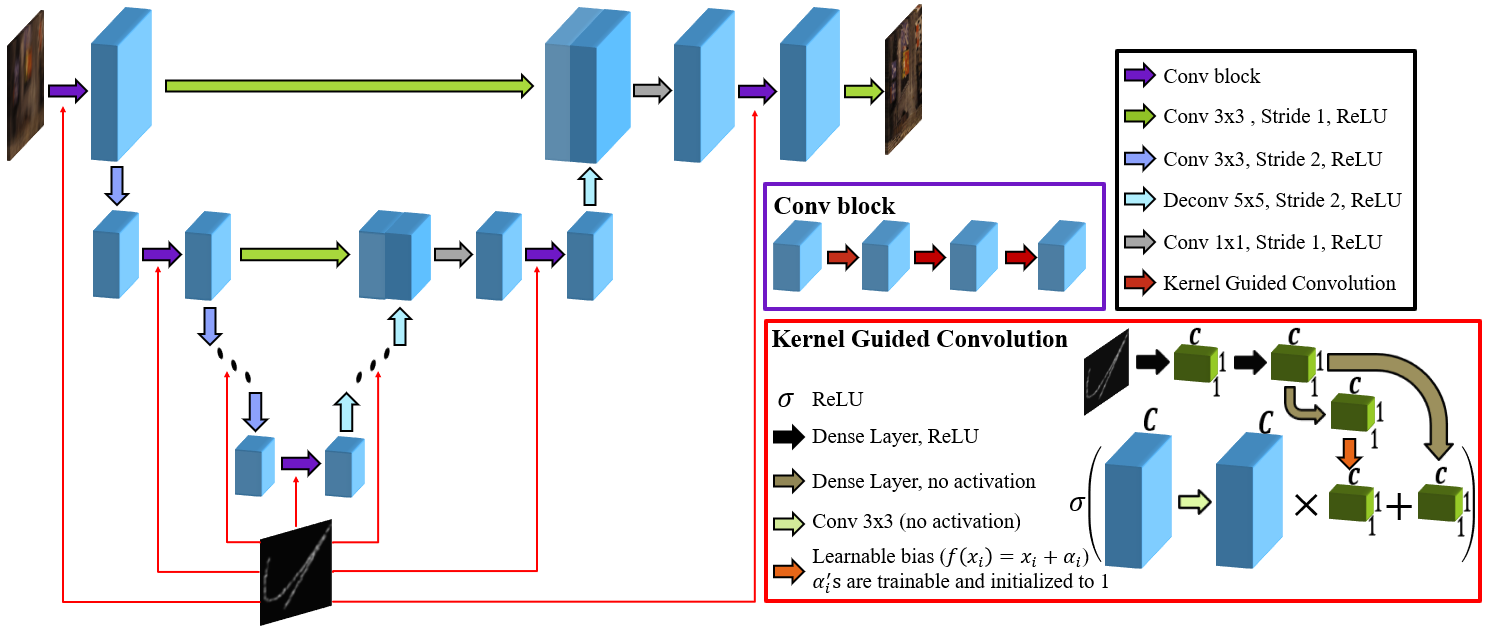}

  \caption{Synthesis Network Architecture. A standard U-Net is augmented with kernel guided convolutions at all its layers. As shown in the red schematic, the activations in these convolutions are modulated and biased by vectors derived from the blur-kernel using FC layers. Each convolution layer consists of 3 successive convolutions using 128 filters, separated by ReLU activations.}
\label{fig:synthesis_arch}
\end{figure*}

\subsection{Synthesis Network}
\label{sec:synthesis}

A uniform 2D blur process is expressed via a convolution, $ B = I*k $, where $B$ is the blurry image, $I$ the blur-free image that we wish to recover, and $k$ is the blur-kernel. In this case, non-blind deblurring boils down to a deconvolution of $B$ with $k$, i.e., a convolution with $k^{-1}$. While the blur-kernel is typically compact, its inverse is generally not compact. Hence, the deconvolution operator is a global operator that operates the same across the image (translation-invariant). This implies that the architecture of deconvolution network should allow such global dependencies as well as a sufficient level of translation-invariance. In these respects, the spatial shrinking and expansion of convolutional auto-encoders~\cite{LeCun89} or U-Nets architecture~\cite{Ronneberger15} fit the bill. Indeed, the use of such architectures for non-blind deblurring is used in~\cite{Nah17,Nimisha17}. It is important to note that a typical blur-kernel is either non-invertable, due to singularities in its spectrum, or close to being so by falling below the noise level. Hence, state-of-the-art non-blind deblurring algorithms incorporate various image priors to cope with the lost data, and the deconvolution networks~\cite{Xu14DNN} are expected to employ their non-linearity to express an operator beyond a linear deconvolution.

In view of these considerations, we opt for using a U-Net as part of the synthesis network that synthesizes the blur-free image given the blurry input. However, global and stationary operation are not the only requirements the discussion above suggests. Being its inverse, the deconvolution kernel $k^{-1}$'s action is completely defined by the blur-kernel $k$ itself, hence the U-Net's action must be dictated by the blur-kernel estimated by the analysis network. Thus, an additional component must be introduced to the synthesis network to accommodate this form of guided non-blind deblurring.

\bf{Kernel Guided Convolution. }\normalfont 
The option of defining the U-net's filter weights as function of the blur-kernel will require a large number of fairly large fully-connected layers in order to model the non-trivial mapping between $k$, which contains thousands of pixels, and the even larger number of weights in the U-Net. Moreover, this additional dependency between the learned weights will increase the degree of non-linearity and may undermine the learning optimization.

A considerably simpler approach is to map the blur-kernel $k$ into a list of biases and multipliers that modulate and shift the output of the convolutions at each layer of the U-Net. More specifically, at each layer of the U-Net the activations from other layers are concatenated together and the first operation they undergo is a convolution. Let us denote the result of this convolution (at a particular layer) by $r$. At this very step, we augment the synthesis network with two vectors, namely the multipliers $m(k)$ and biases $b(k)$, that depend on the blur-kernel $k$ and operate on $r$ by
\begin{equation}
\label{eq:unit}
r = r \odot (1+m(k))+b(k),
\end{equation}
where $\odot$ denotes a point-wise multiplication. As prescribed by the discussion above, this operation allows the blur-kernel to affect the U-Net's action uniformly across the entire spatial domain. 

The synthesis-guiding unit, that models the functional dependency of $m(k) $ and $ b(k)$ over the blur-kernel, is depicted in Figure~\ref{fig:synthesis_arch} and consists of three layers of fully-connected layers, biases and ReLU operations (except for the last layer). The input dimension is the total number of pixels in the blur-kernel, the intermediate dimension is set to 128, and the final dimensions equals to the number of channels at that layer. Each layer of the U-Net is augmented with it own synthesis-guiding unit, as shown in Figure~\ref{fig:synthesis_arch}. We note that a similar form of control units was recently suggested in the context of image translation~\cite{Huang18}. In Section~\ref{sec:results} we report an ablation study that demonstrates the benefit in incorporating the guided convolution.

\emph{Architecture Details.} The U-Net architecture we use is the convolutional encoder-decoder one described in~\cite{Ronneberger15} and depicted in Figure~\ref{fig:synthesis_arch}. This encoder-decoder architecture has connections between every layer in the encoder and the decoder at the same depth. The number of channels at all the layers is set to 128. The un-pooling in the decoder is followed by a convolution with 128 filters of size 5-by-5 pixels and the ones arriving from the corresponding encoding are convolved with the same number of filters whose size is 3-by-3. These activations are concatenated into a 256 channels tensor which is reduced back to 128 channels by 128 filters of size 1-by-1 pixels. These activations as well as the ones in the encoding layers go through the $k$-dependent multiplicative and additive steps in Eq.~\ref{eq:unit} which are produced by layer-specific synthesis-guiding units. The resulting activations go through three steps of convolution with 128 filters of size 3-by-3 interleaved with ReLU operation as shown in Figure~\ref{fig:synthesis_arch}. The pooling and un-pooling steps between the layers perform x2 scaling. Finally, the synthesis network is trained to minimize $L_2$ reconstruction loss with respect to the corresponding sharp training images

\begin{figure}[t]
\begin{center}

\rotatebox{90}{Analysis\hspace{0.9em} E2E\hspace{2em} GT}
\includegraphics[width=0.152\linewidth]{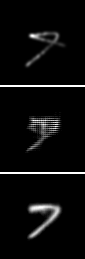} 
\hspace{-1mm}
\includegraphics[width=0.152\linewidth]{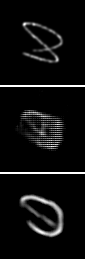}
\hspace{-1mm}
\includegraphics[width=0.152\linewidth]{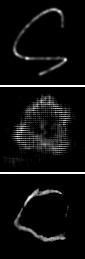} 
\hspace{-1mm}
\includegraphics[width=0.152\linewidth]{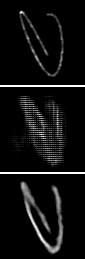}
\hspace{-1mm}
\includegraphics[width=0.152\linewidth]{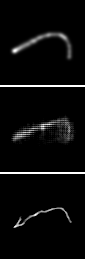}
\hspace{-1mm}
\includegraphics[width=0.152\linewidth]{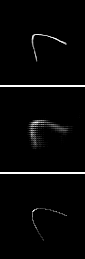} 

\end{center}
  \caption{Estimated Kernels. We show here the kernel estimated by the analysis network once it was pre-trained to estimate the ground-truth kernel, and once it was further trained to minimize the image loss in the end-to-end training step.}
\label{fig:kernel_comp}
\end{figure}

\subsection{Scale Optimized Networks}
\label{sec:multinet}
As we report in Section~\ref{sec:results}, a further increase in deblurring accuracy is obtained by allowing the analysis and synthesis networks to specialize on a narrower range of blur-kernel sizes. Specifically, we split the kernel size range into three segments $[0,31],(31,61]$ and $(61,85]$ and train a network to classify an input blurry image into one of these three classes of kernel size. We then use the classification of this network to train three pairs of analysis and synthesis network, each trained over images belonging to the different class of blur size. As a classification network we used a single scale analysis network augmented with an output fully-connected layer with three neurons and trained it using a categorical cross entropy loss.


\subsection{Training}
\label{sec:training}

There are several options for training the analysis and synthesis networks. One possible approach is to train both networks to optimize the final sharp image reconstruction loss, which is our ultimate goal. Our experiments however show that better convergence is attained using the following training strategy.

Both the analysis and synthesis networks are first pre-trained independently to optimize the kernel and image-reconstruction losses respectively. Unlike in other approach, this stage makes use of the ground-truth blur-kernels. We then train the entire network, i.e., the composition of the analysis and synthesize networks, to optimize the output image reconstruction while \emph{omitting} blur-kernel reconstruction loss.

This end-to-end (E2E) training allows the synthesis network to learn to cope with the inaccuracies in the estimated kernel, as well as trains the analysis network to estimates the blur under the metric induced by the image reconstruction loss. Furthermore, as shown in Figure~\ref{fig:kernel_comp}, omitting the kernel loss allows the networks to abstract its representation for achieving a better overall deblurring performance. In Section~\ref{sec:results} we report an ablation study exploring the benefit in our training strategy versus alternative options.

\section{Results}
\label{sec:results}

\emph{Implementation.} We implemented the networks using the Keras library~\cite{keras15} and trained them using two Geforce GTX 1080 Ti GPUs running in parallel. The training was performed using an ADAM optimizer~\cite{Adam14} with an initial learning-rate of $10^{-4}$ which was reduced by $20\%$ whenever the loss stagnated over 5 epochs. We used a small batch size of 4 examples. This required us to perform a fairly large number of training iterations, around $10^6$ at the pre-training stage, and another $10^5$ iterations at the end-to-end stage. At run time, its takes our trained network about 0.45 seconds to deblur a 1024-by-768 pixels image.


\begin{table}[t]
\hspace{-0.1in}
\begin{tabular}{l|cc|cc}

\multirow{2}{*}{\textbf{Method}} & \multicolumn{2}{c|}{All Trajectories}  & \multicolumn{2}{c}{Excluding 8,9,10} \\
 & \textbf{PSNR} & \! \!\!\! \textbf{MSSIM} & \textbf{PSNR} & \!\!\!\! \textbf{MSSIM}\\
\hline
\textbf{Deblurring Algos.}  &   & &   & \\

Cho~\cite{Cho09}& 28.98& 0.933
& 30.39 & 0.953\\

Xu~\cite{Xu10}& 29.53& \textbf{0.944}& 31.05 & \textbf{0.964}\\

Shan~\cite{Shan08}& 25.89&0.842 & 27.57 & 0.905\\
Fergus~\cite{Fergus06}& 22.73 &0.682& 24.06 &0.756 \\

Krishnan~\cite{Krishnan11}& 25.72 &0.846& 27.90 & 0.909\\

Whyte~\cite{Whyte11}& 28.07&0.848 & 30.82 &0.941 \\

Hirsch~\cite{Hirsch11}& 27.77 &0.852& 30.01 & 0.945\\

\hline
\textbf{Deblurring Nets.}  &   & &   & \\

DeepDeblur~\cite{Nah17}& 26.48 &0.807& - &- \\

DeblurGAN~\cite{Kupyn17}&26.10 &0.816&- &- \\

DeblurGAN-v2\cite{Kupyn19}&26.97 &0.830&28.87 &0.921 \\

SRN~\cite{Tao18}&27.06 &0.840& 29.22 &0.923 \\

Ours& 29.97&0.915 & 31.87 &0.961 \\
Ours scale opt. nets& \bf{30.22}&0.915 & \bf{32.24} &0.961 \\
\end{tabular}
\vspace{0.08in}
\caption{Table reports the mean PSNR and MSSIM obtained over the K\"ohler dataset~\cite{Kohler12}. 
Note that DeepDeblur~\cite{Nah17}, DeblurGAN~\cite{Kupyn17}, DeblurGAN-v2~\cite{Kupyn19} and SRN~\cite{Tao18} are network based methods, whereas the rest are blind deblurring algorithms. The scores were computed using the script provided in~\cite{Kohler12}. The output images of DeepDelur and DeblurGAN were not available for us for calculating the scores without the large trajectories.}
\label{tab:kohler_numeric}
\end{table}

\emph{Training Set.} We obtained the sharp images from the Open Image Dataset~\cite{openimage18}, and used random crops of 512-by-512 pixels to produce the training samples. Each training example is produced on the fly and consists of a triplet of a sharp image sample $I$, a blur-kernel $k$, and the resulting blurred image $B=I*k+\eta$, where $\eta\sim\mathcal{N}(0,0.02^2)$. As noted in Section~\ref{sec:analysis}, the analysis network operates only on Y channel of the input image. Normalizing this channel to have a zero mean and unit variance does not effect the blur, but appears to improve the convergence of its estimation.

The training set was produced from 42k images, and the test set was produced from a separate set of 3k images. The blur-kernels we used for simulating real-world camera shake were produced by adapting the method in~\cite{Chakrabarti16}. This consists of randomly sampling a spline, mimicking camera motion speed and acceleration, as well as a various sensor point spread functions. We make this code available along with the implementation of our networks.

Finally, let us note that our networks were trained once, over this training set, and were evaluated on test images arriving from other benchmark datasets. 

\begin{table}[t]
\begin{center}

\begin{tabular}{lcc}
\textbf{Method} & \textbf{PSNR} & \textbf{MSSIM}\\
\hline
\textbf{Deblurring Algos.}  &   &\\
Sun~\cite{Sun13}& 20.47 & 0.781\\
Zhong~\cite{Zhong13}& 18.95 & 0.719 \\
Fergus~\cite{Fergus06}& 15.60 &0.508 \\
Cho~\cite{Cho09}& 17.56 & 0.636 \\
Xu~\cite{Xu10}& 20.78 & 0.804\\
Krishnan~\cite{Krishnan11}& 17.64 & 0.659\\
Levin~\cite{Levin11}& 16.57 & 0.569\\
Whyte~\cite{whyte12}& 17.444 & 0.570 \\
Xu~\cite{Xu13}& 19.86 & 0.772\\
Zhang~\cite{Zhang13}& 16.70 & 0.566 \\
Michaeli~\cite{Michaeli14}&18.91 &0.662 \\
Pan~\cite{Pan14} & 19.33 & 0.755\\
Perrone~\cite{Perrone14} & 19.18& 0.759\\
\hline
\textbf{Deblurring Nets.}  &   &\\
DeblurGAN-v2~\cite{Kupyn19}&17.98 & 0.595 \\
SRN~\cite{Tao18}&17.28 & 0.590 \\

Ours & 20.67&0.799 \\
Ours scale optimized& \textbf{20.89}&\textbf{0.819}
\end{tabular}

\end{center}
\caption{
Mean PSNR and SSIM on the Lai synthetic uniform-blur dataset~\cite{Lai16}. DeblurGAN-v2~\cite{Kupyn19} and SRN~\cite{Tao18} are deep learning methods, whereas the other are classical blind deblurring methods. The scores were computed by adapting the script provided in~\cite{Kohler12}
}
\label{tab:lai_numeric}
\end{table}

\bf{K\"ohler Dataset. }\normalfont This dataset~\cite{Kohler12} consists of 48 blurred images captured by a robotic platform that mimics the motion of a human arm. While the blur results from 6D camera motion, most of the images exhibit blur that ranges from being approximately uniform up to admitting a 3D blur model. Trajectories 8,9, and 10 in this set contain blur which is larger than the maximal kernel size we used in our training ($m=85$). Hence, on these specific trajectories, we first applied an x2 downscaling, applied our networks, and upsampled the result back to the original resolution. This interpolation reduced the reconstruction accuracy by 1.6dB on average, and can be avoided by training a larger network.

\begin{figure*}[!htb]

\begin{tabular}
 {p{0.22\linewidth}p{0.22\linewidth}p{0.22\linewidth }p{0.22\linewidth }}
\begin{center}\textbf{Input} \end{center} & 
\begin{center}\textbf{SRN}~\cite{Tao18} \end{center} & 
\begin{center}\textbf{DeblurGAN-v2}~\cite{Kupyn19} \end{center} & 
\begin{center}\textbf{Ours} \end{center}  
\end{tabular}

\includegraphics[width=1\linewidth]{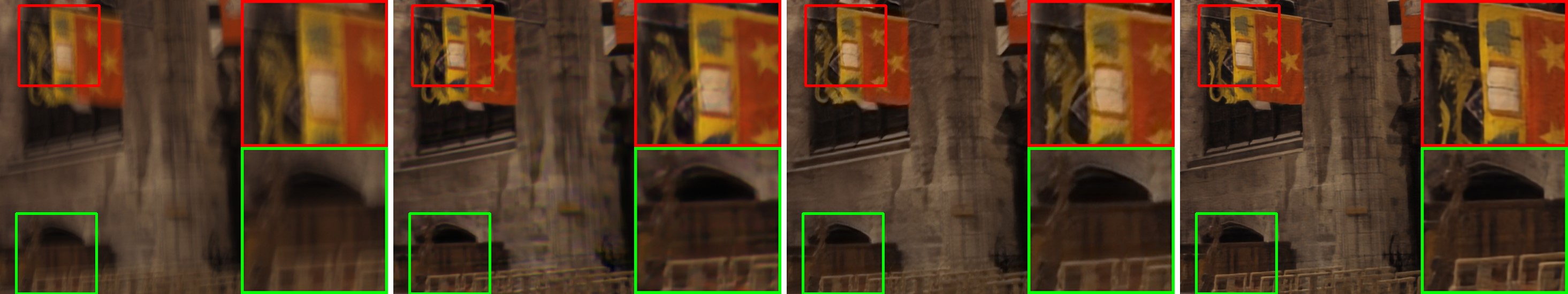}


\vspace{1mm}
\includegraphics[width=1\linewidth]{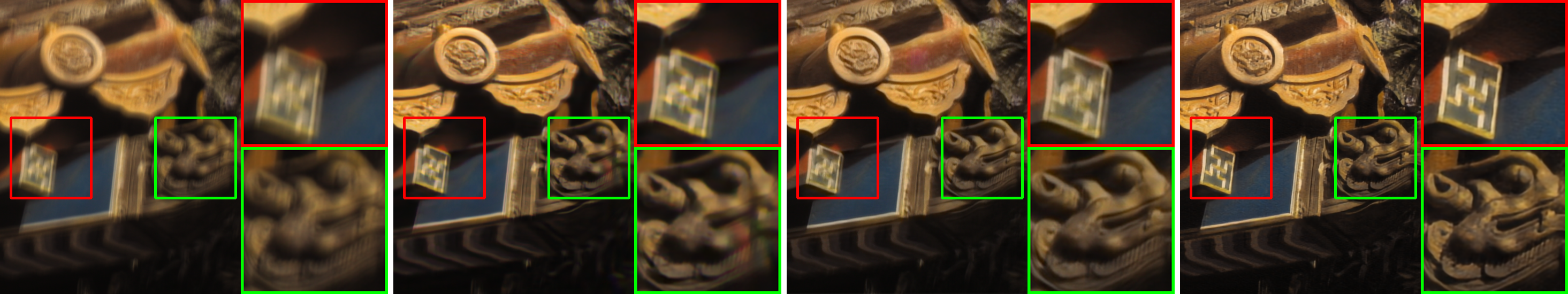}

  \caption{Visual comparison against alternative deblurring networks, over the K\"ohler dataset.}
\label{fig:kohler_visual}

\end{figure*}

As Table~\ref{tab:kohler_numeric} shows, while our network is in par with state-of-the-art deblurring algorithms, with and without taking into account trajectories 8,9, and 10, it achieves a speedup of x10 and higher in deblurring time. Compared to state-of-the-art deblurring networks, Table~\ref{tab:kohler_numeric} shows a significant improvement of 3.16dB in deblurring accuracy, which we attribute to our novel design of analysis and synthesis networks pair which explicitly estimate the blur-kernel and use it to guide the deblurred-image synthesis. Figure~\ref{fig:kohler_visual} shows some of the images produced by the SRN~\cite{Tao18}, DelburGAN-v2~\cite{Kupyn19} and our networks.

\begin{figure}

\begin{tabular}{p{0.18\linewidth}p{0.22\linewidth}p{0.2\linewidth}p{0.2\linewidth}}
\centering \textbf{Input}& 
\centering\textbf{DeblurGAN-v2}\cite{Kupyn19}& 
\centering\textbf{SRN}\cite{Tao18}& 
\centering\textbf{Ours}
\end{tabular}
\includegraphics[width=1\linewidth]{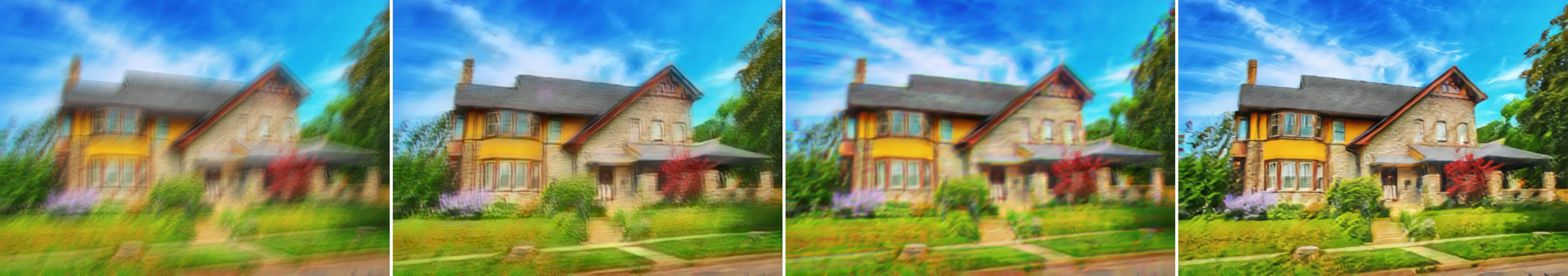}\\
\includegraphics[width=1\linewidth]{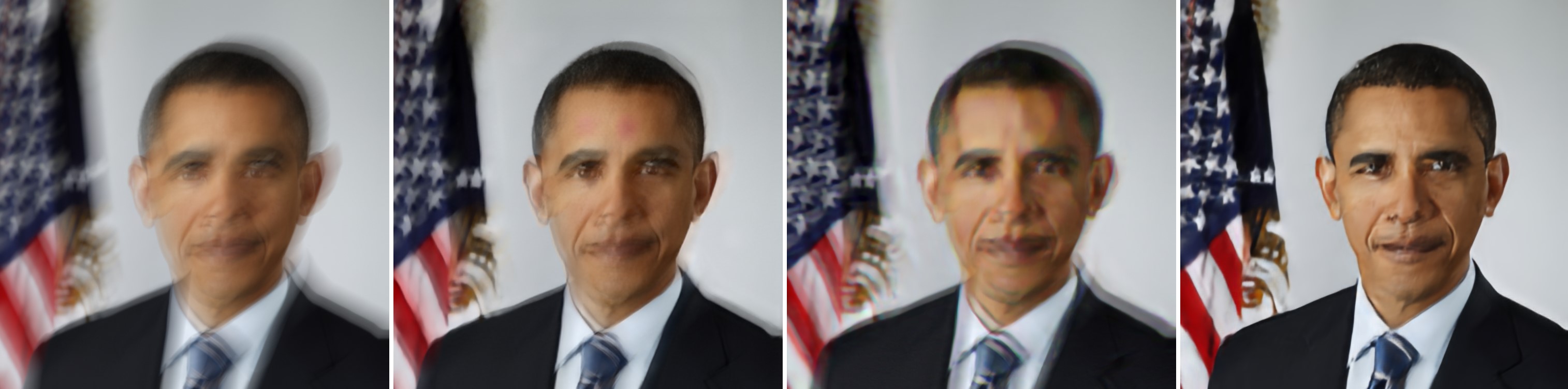}\\
\includegraphics[width=1\linewidth]{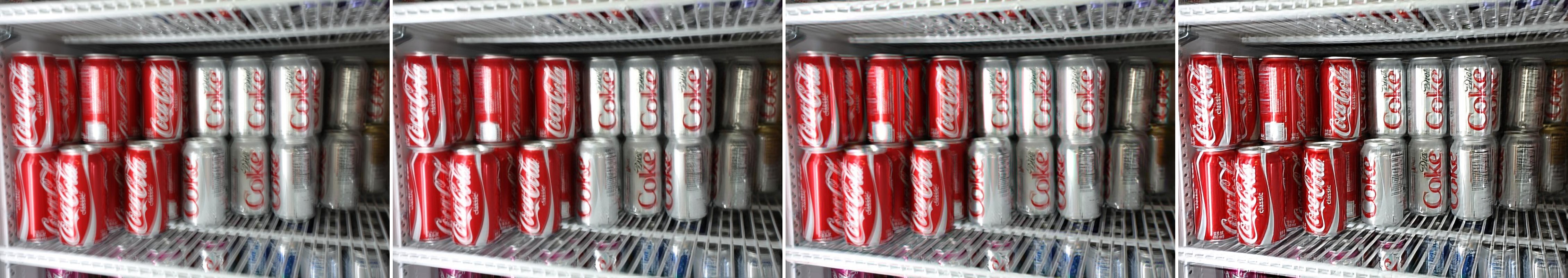}\\
\includegraphics[width=1\linewidth]{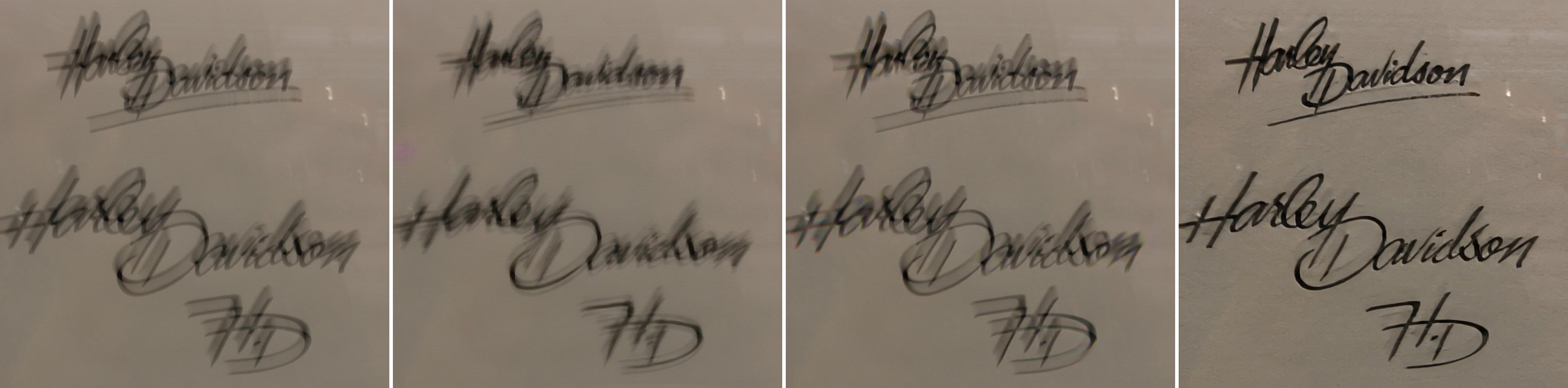}

  \caption{Visual comparison on blurry images from Lai dataset. The top two images are synthetically blurred, while the bottom two are real-world blurry images.}
\label{fig:lai_visual}
\end{figure}


  




\bf{Lai Dataset. }\normalfont This dataset~\cite{Lai16} consists of real and synthetic blurry images. The synthetic dataset is created by both uniform and non-uniform blurs. We focus on the former, which was produced by blurring 25 high-definition images with 4 kernels. This dataset is more challenging due to the strong blur and saturated pixels it contains.

Table~\ref{tab:lai_numeric} portrays a similar picture, where where our method achieves state-of-the-art deblurring accuracy, and outperforms existing deblurring networks. Finally, Figure~\ref{fig:lai_visual} provide a visual comparison between the deblurring networks and ours, on synthetic and real-world images respectively, both taken from this dataset. Again, our resulting deblurred images appear sharper where some of the letters are made more readable.

\emph{Ablation study.} Finally, we analyze the contribution of different elements in our design, and report in Table~\ref{tab:ablation_study} the mean PSNR they produce over our validation set. The table reports the deblurring accuracy obtained by the synthesis network with different configurations of the guided convolution as well as the inferior results obtained with without it (loss of 3.93dB). The table also reports the added value of our end-to-end training, where the analysis network's output is plugged into the synthesis network during training (gain of 1.96dB). The importance of initiating this training stage with pre-trained analysis and synthesis networks is also evident from the table (gain of 1dB). Finally, A small improvement can be achieved by using multiple networks, as described in \ref{sec:multinet}.

\begin{table}[]
\begin{center}

\begin{tabular}{lc}
\textbf{Configuration} & \textbf{PSNR}\\
\hline
\textbf{Synthesis + GT Kernels}  &  \\
No guidance &24.80 \\
Additive guidance&28.58\\
Multiplicative guidance&28.41\\
Additive+Multiplicative guidance&28.73\\
\hline
\textbf{Training Strategy}  &  \\
Random initialization & 25.70 \\
Pre-training before E2E training& 24.82 \\
Pre-training after E2E training & 26.78 \\
scale opt. nets & 27.12 

\end{tabular}
\end{center}
\caption{Ablation study. Table reports the mean PSNR obtained by different elements of our design. Our validation set is used for this evaluation. The guided convolution is explored in different settings in which the multiplication and additive terms are omitted from Eq.~\ref{eq:unit}.}
\label{tab:ablation_study}
\end{table}

\section{Discussion}
\label{sec:discussion}

We presented a new deblurring network which, like existing deblurring algorithms, breaks the blind image deblurring problem into two steps. Each step is tackled with a dedicated and novel task-specific network architecture. This design allows us to explicitly make use of the blur-kernel during the networks' training. The analysis network contains novel correlation layers allowing it to mimic and generalize successful kernel-estimation procedures. The synthesis network contains unique convolution guided layers that allow the estimated kernel to control its operation.


While we derived and implemented our approach for 2D uniform blur, we believe that can be extended to 3D rotational blur, which will require adding a third axis of rotation into the cross-correlation layers. 


{\small
\bibliographystyle{ieee_fullname}
\bibliography{egbib}
}

\end{document}